\documentclass{article}
\usepackage{ais25}
\usepackage[T1]{fontenc}
\usepackage[table, dvipsnames]{xcolor}
\usepackage{amsmath}
\usepackage{amsfonts}

\usepackage{hyperref}
\usepackage{url}
\hypersetup{
    citecolor=CadetBlue,
    colorlinks=true,
    linkcolor=CadetBlue,
    filecolor=magenta,      
    urlcolor=cyan}

\newcommand{\model}[1]{\texttt{#1}}

\usepackage{mathtools}
\DeclarePairedDelimiter\customnorm{\lVert}{\rVert}

\title{Towards Foundation Inference Models that Learn ODEs In-Context}
\author{\name{Maximilian Mauel$^2$}, {Manuel Hinz$^{12}$}, {Patrick Seifner$^{12}$}, \\ {David Berghaus$^{13}$} \& {Rams\'es J. S\'anchez$^{123}$}  \\ \addr{$^1$Lamarr Institute, $^2$University of Bonn, $^3$Fraunhofer IAIS} \\ \email{seifner@cs.uni-bonn.de}}

\begin{document}
\maketitle

\begin{abstract}
Ordinary differential equations (ODEs) describe dynamical systems evolving deterministically in continuous time. 
Accurate data-driven modeling of systems as ODEs, a central problem across the natural sciences, remains challenging, especially if the data is sparse or noisy. 
We introduce \model{FIM-ODE} (Foundation Inference Model for ODEs), a pretrained neural model designed to estimate ODEs \textit{zero-shot} (i.e.~\textit{in-context}) from sparse and noisy  observations. 
Trained on synthetic data, the model utilizes a flexible neural operator for robust ODE inference, even from corrupted data. 
We empirically verify that \model{FIM-ODE} provides accurate estimates, on par with a neural state-of-the-art method, and qualitatively compare the structure of their estimated vector fields. 
\end{abstract}

\section{Introduction}

Ordinary differential equations (ODEs) are a class of continuous-time dynamical systems, modeling phenomena evolving deterministically in continuous space. 
They are formally described by \textit{vector fields} $\mathbf{f}: \mathbb{R}_+ \times \mathbb{R}^D \rightarrow \mathbb{R}^D$, defining the differential equation 
\begin{equation}
    d \mathbf{x}(t) = \mathbf{f}(t, \mathbf{x}(t)) \, dt,  
\end{equation}
with continuous solution trajectories $\mathbf{x}: \mathbb{R}_+\rightarrow \mathbb{R}^D$. 
The ubiquitous presence of these simple models throughout the natural sciences is remarkable, from describing Newton's laws of motion, to population dynamics in biological systems \citep{lotka-1925-elements,volterra-1926-variazioni} and atmospheric convection in meteorology \citep{lorenz-1963-atmospheric}. 
Accurate ODE models characterize the underlying phenomena (e.g. by fixed points or limit cycles) and can forecast future states. 

In this work, we consider the \textit{ODE system identification problem}: estimating the ODE (i.e. a vector field) that best describes a system from time series observations only. 
Traditional approaches to this problem are either \textit{non-parametric}
\citep{heinonen-2018-node} or \textit{symbolic-regression}-based \citep{dong-2025-symbreg}. 
Recently, \citet{dascoli-2024-odeformer} introduced \model{ODEFormer}, a pretrained \textit{neural symbolic regression} method, transforming (tokenized) time series observations into (tokenized) equations. 

Foundation Inference Models (FIMs) are a general simulation-based framework for inference of dynamical systems.  
These deep neural network models are pretrained on synthetic data, generated from a class of dynamical systems, using a \textit{supervised} train objective to \textit{match} the target and estimated infinitesimal generators of the processes.
This framework has shown promising results in the inference of continuous-time Markov chains \citep{berghaus-2024-mjp}, stochastic differential equations \citep{seifner-2025-sde}, and point processes \citep{berghaus-2025-pp}. 
If appropriate, it defaults to \textit{neural operators} \citep{kovachki-2023-operators} for continuous function estimates \citep{seifner-2025-imputation}. 

In this work, we introduce \model{FIM-ODE}, a FIM for the ODE identification problem, which uses a \textit{neural operator} to model the vector field. 
Our model performs better than \model{ODEFormer} on a synthetic test set and we argue that \model{FIM-ODE}'s flexible neural operator yields more reasonable global vector field predictions.

\section{Foundation Inference Model}
The ODE inference problem is a special case of the stochastic differential equation (SDE) inference problem, tackled by \model{FIM-SDE} \citep{seifner-2025-sde}. 
However, non-stochastic trajectories \textit{do not explore the space as much as stochastic trajectories}, providing less information about the global vector field. 
Nevertheless, because of the problem similarities, we stick to the architecture of \model{FIM-SDE}.

\textbf{Synthetic Data Generation.}
Following \citet{seifner-2025-sde}, we generate synthetic training data from a broad distribution over ODE systems and initial states of up to dimension $D=3$. 
Each component of a vector field is sampled from a distribution over multivariate polynomials of up to degree $3$. 
We simulate these systems, record the solutions at (irregular) grids and corrupt them by additive noise to generate time series observations from them. 

\textbf{Inference Model Architecture.}
\model{FIM-ODE} estimates a vector field $\mathbf{\hat{f}}$ from a set $\mathcal{D} = \{\mathbf{y}_{k}\}_{k=1}^K$ of time series  $\mathbf{y}_k = [(t_{k1}, \mathbf{y}_{k1}), \dots, (t_{kL}, \mathbf{y}_{kL})]$ \textit{in-context}, without any further training or finetuning. 
The vector field estimate $\mathbf{\hat{f}}$ is implemented by a \textit{DeepONet} neural operator \citep{lu-2021-deeponet}. 
The \textit{branch-net} of \model{FIM-ODE}, a Transformer encoder, encodes $\mathcal{D}$ into $K(L-1)$ $E$-dimensional representations $\mathbf{D} \in \mathbb{R}^{E\times K(L-1)}$. 
Note that this retains individual encodings for almost all observations, which a \textit{combination network} (see below) can access directly. 
The \textit{trunk-net} is a linear map, encoding a location $\mathbf{x}\in \mathbb{R}^D$ into $\mathbf{h}(\mathbf{x})\in \mathbb{R}^E$. 
The \textit{combination network} is a sequence of residual attention layers, using $\mathbf{D}$ as keys and values, and $\mathbf{h}(\mathbf{x})$ as query, similar to a Transformer decoder. 
A final linear projection yields the estimate $\mathbf{\hat{f}}(\mathbf{x})$.

\textbf{Training.}
Adhering to the FIM framework, we use a \textit{supervised training objective} $\mathcal{L}(\mathbf{x}, \mathcal{D}, \mathbf{f})  = \customnorm{\mathbf{\hat{f}} (\mathbf{x}) - \mathbf{f}(\mathbf{x})}^2$, matching the predicted to the ground-truth vector fields on $\mathbf{x}$ sampled ``close'' to the observation values in $\mathcal{D}$. 
We pretrain a single \model{FIM-ODE} with roughly $20$M parameters on $600$k synthetic equations\footnote{For comparison, \model{ODEFormer} is trained on $50$M equations and has $86$M parameters.}.

\section{Experiments}
For our experiments, we sample $4000$ ODEs with polynomial vector fields of up to degree $3$ and up to dimension $3$. 
We simulate $9$ trajectories with initial states sampled from $\mathcal{N}(0,1)$ from each ODE, recording $200$ observations on a regular grid with inter-observation times $\Delta \tau = 0.05$. 
Following \citet{dascoli-2024-odeformer}, we measure performance by the \textit{$R^2$-accuracy}, the percentage of $R^2$ scores larger than $0.9$. 
We compare our pretrained \model{FIM-ODE} to a pretrained \model{ODEFormer}\footnote{Source of pretrained \model{ODEFormer} weights: \url{https://github.com/sdascoli/odeformer}} on two tasks: 
a \textit{reconstruction task}, in which performance is measured on the reconstruction of the \textit{context trajectory}, and a \textit{generalization task}, in which performance is measured on the reconstruction of \textit{held-out trajectories}. 
Note that \model{ODEFormer} can only process a single context trajectory, so we restrict to that case in direct comparisons. 

\begin{table}
\caption{Comparison of $R^2$-accuracy on polynomial vector field data, given a single trajectory as context.}
\begin{center}
\begin{tabular}{@{}lcc@{}}
\toprule
Model & Reconstruction Task & Generalization Task \\
    \midrule
\model{ODEFormer} & $0.65$ & $0.19$ \\
\model{FIM-ODE} & $\mathbf{0.90}$ &  $\mathbf{0.26}$ \\
    \bottomrule
\end{tabular}
\label{tab:reconst-general}
\end{center}
\end{table}

\textbf{Reconstruction and Generalization Performance.} 
Table~\ref{tab:reconst-general} contains the $R^2$-accuracy for both tasks on the polynomial test set. 
Note that the polynomial test data is in the training distribution of both models, but \model{ODEFormer} was trained on a broader distribution than \model{FIM-ODE}. 
Our model performs well in both tasks, even better than \model{ODEFormer}.  
The large difference is $R^2$-accuracy shows that generalization is inherently more difficult than reconstruction. 

\begin{figure}[t]
\begin{center}
    \includegraphics[width=0.85\textwidth]{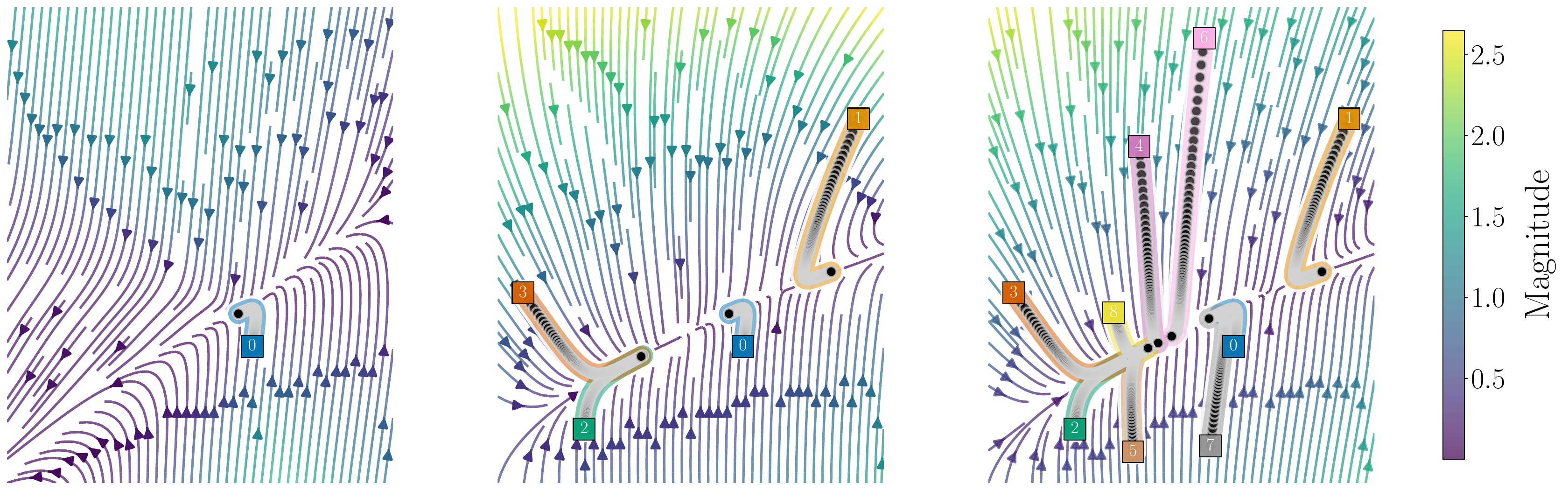}
\end{center}
\caption{
{
Vector field estimates of \model{FIM-ODE} from $1$, $5$ and $9$ context trajectories of the same system. 
The estimates improve, given more trajectories. 
}
}
\label{fig:fim-num-traj}
\end{figure}

\textbf{Multi-trajectory Context.}
\model{FIM-ODE} is designed to process multiple trajectories of the same system simultaneously. 
Figure~\ref{fig:fim-num-traj} depicts vector field estimates of \model{FIM-ODE} with varying counts of context trajectories. 
For a single trajectory, vector field estimates are inaccurate at locations distant from the observations.
With more trajectories covering the space, \model{FIM-ODE} corrects these estimates, extracting and combining all available information effectively. 
Such accurate global vector field prediction is particularly beneficial for generalization tasks --- as well as for \textit{interpretability in scientific settings}.

\begin{figure}[t]
\begin{center}
    \includegraphics[width=0.85\textwidth]{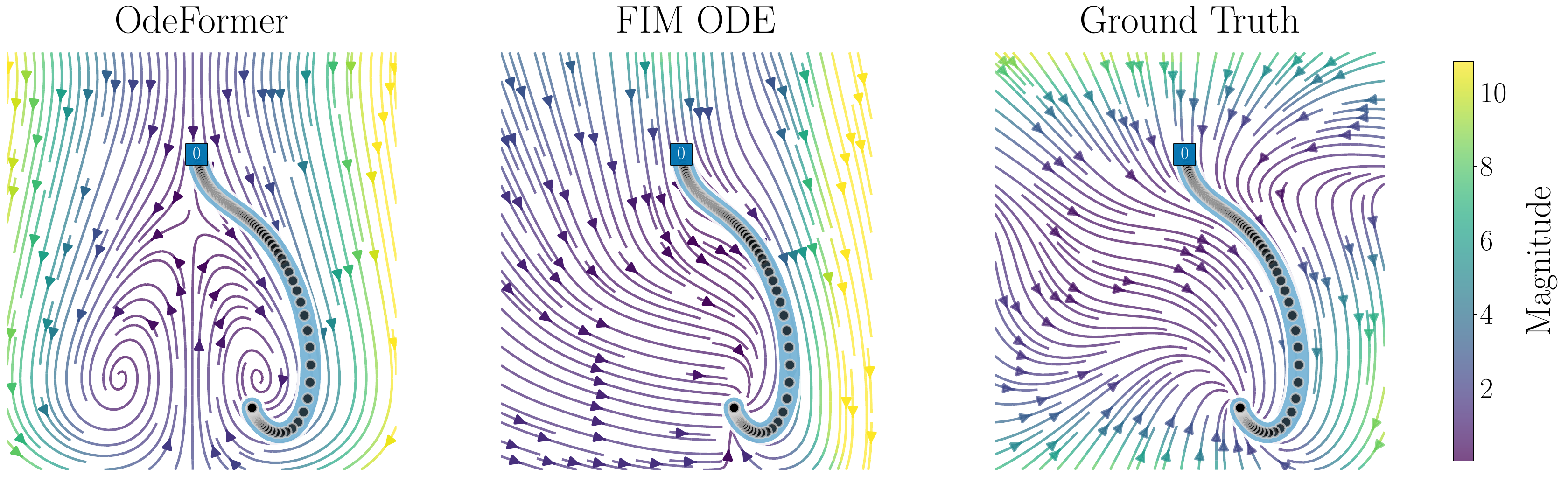}
\end{center}
\caption{
Comparison of vector field estimates to the ground-truth, given a single trajectory. 
\model{ODEFormer} predicts a complex global pattern, while \model{FIM-ODE} focuses on accurate predictions near the provided trajectory. 
}
\label{fig:fim-vector-field-to-odeformer}
\end{figure}

\textbf{Local and Global Predictions.} 
In Figure~\ref{fig:fim-vector-field-to-odeformer}, we visually compare vector field estimates of \model{FIM-ODE} and \model{ODEFormer} from a single trajectory.
Locally, at the observations, \model{FIM-ODE} predicts a complex pattern to reconstruct the trajectory. 
Globally, far away from the observations, the predictions are much simpler. 
Compared to the ground-truth, \model{ODEFormer} predicts a more complex vector field. 
This results in complex global pattern predictions, largely unjustified by the single, quite simple observed trajectory. 
The stark difference between the two models is explained by the different vector field parametrizations. 
\model{ODEFormer} is restricted to predict (rational) polynomial symbolic equations, which, given sparse or noisy observations, might not default to simple expressions. 
\textit{The neural operator of \model{FIM-ODE} can handle these situations more flexible}.

\section{Conclusions}
In this work, we introduced \model{FIM-ODE}, a pretrained foundation inference model that estimates ODEs from time series observations, utilizing a neural operator. 
In our preliminary experiments, it demonstrates better performance than \model{ODEFormer}, a state-of-the-art neural symbolic regression approach. 
We find structural differences between the vector fields predicted by these approaches and argue that \model{FIM-ODE}'s global predictions are more justifiable. 

In future work, we will compare these methods on ODEBench, a ODE inference benchmark dataset of $63$ hand-selected ODEs. 
We will also explore the discovery of latent dynamics using a pretrained \model{FIM-ODE}, with possible applications in neural population and chemical reaction dynamics~\citep{duncker2019learning}, and the evolution of natural language content~\citep{cvejoski2023neural, cvejoski2022future}. 
\section*{Acknowledgments}

This research has been funded by the Federal Ministry of Education and Research of Germany and the state of North-Rhine Westphalia as part of the Lamarr Institute for Machine Learning and Artificial Intelligence.

\bibliography{ais}

\end{document}